\documentclass[final]{cvpr}

\usepackage{times}
\usepackage{epsfig}
\usepackage{graphicx}
\usepackage{amsmath}
\usepackage{amssymb}
\usepackage{multirow}
\usepackage{color}

\def\confYear{CVPR}

\begin{document}

\title{Beyond Self-Supervision: A Simple Yet Effective Network Distillation Alternative to Improve Backbones}  
\author{Cheng Cui, Ruoyu Guo, Yuning Du, Dongliang He, Fu Li, Zewu Wu, Qiwen Liu, \\Shilei Wen, Jizhou Huang, Xiaoguang Hu, Dianhai Yu, Errui Ding, Yanjun Ma  \\
Baidu Inc.\\
\tt\small \{cuicheng01, guoruoyu, duyuning, hedongliang01, lifu, wuzewu, liuqiwen,\\ 
\tt\small wenshilei, huangjizhou, huxiaoguang, yudianhai, dingerrui, mayanjun02\} @baidu.com
}
\maketitle
\thispagestyle{empty}

\begin{abstract}

Recently, research efforts have been concentrated on revealing how pre-trained model makes a difference in neural network performance. Self-supervision and semi-supervised learning technologies have been extensively explored by the community and are proven to be of great potential in obtaining a powerful pre-trained model. However, these models require huge training costs (i.e., hundreds of millions of images or training iterations). In this paper, we propose to improve existing baseline networks via knowledge distillation from off-the-shelf pre-trained big powerful models. Different from existing knowledge distillation frameworks which require student model to be consistent with both \emph{soft-label} generated by teacher model and \emph{hard-label} annotated by humans, our solution performs distillation by only driving prediction of the student model consistent with that of the teacher model. Therefore, our distillation setting can get rid of manually labeled data and can be trained with extra unlabeled data to fully exploit capability of teacher model for better learning. We empirically find that such simple distillation settings perform extremely effective, for example, the top-1 accuracy on ImageNet-1k validation set of MobileNetV3-large and ResNet50-D can be significantly improved from 75.2\% to 79\% and 79.1\% to 83\%, respectively. We have also thoroughly analyzed what are dominant factors that affect the distillation performance and how they make a difference. Extensive downstream computer vision tasks, including transfer learning, object detection and semantic segmentation, can significantly benefit from the distilled pretrained models. All our experiments are implemented based on PaddlePaddle$\footnote{https://github.com/PaddlePaddle}$, codes and a series of  improved pretrained models with ssld suffix are available in PaddleClas $\footnote{https://github.com/PaddlePaddle/PaddleClas}$.

\end{abstract}

\begin{figure}[t]
\centering
\includegraphics[width=\columnwidth]{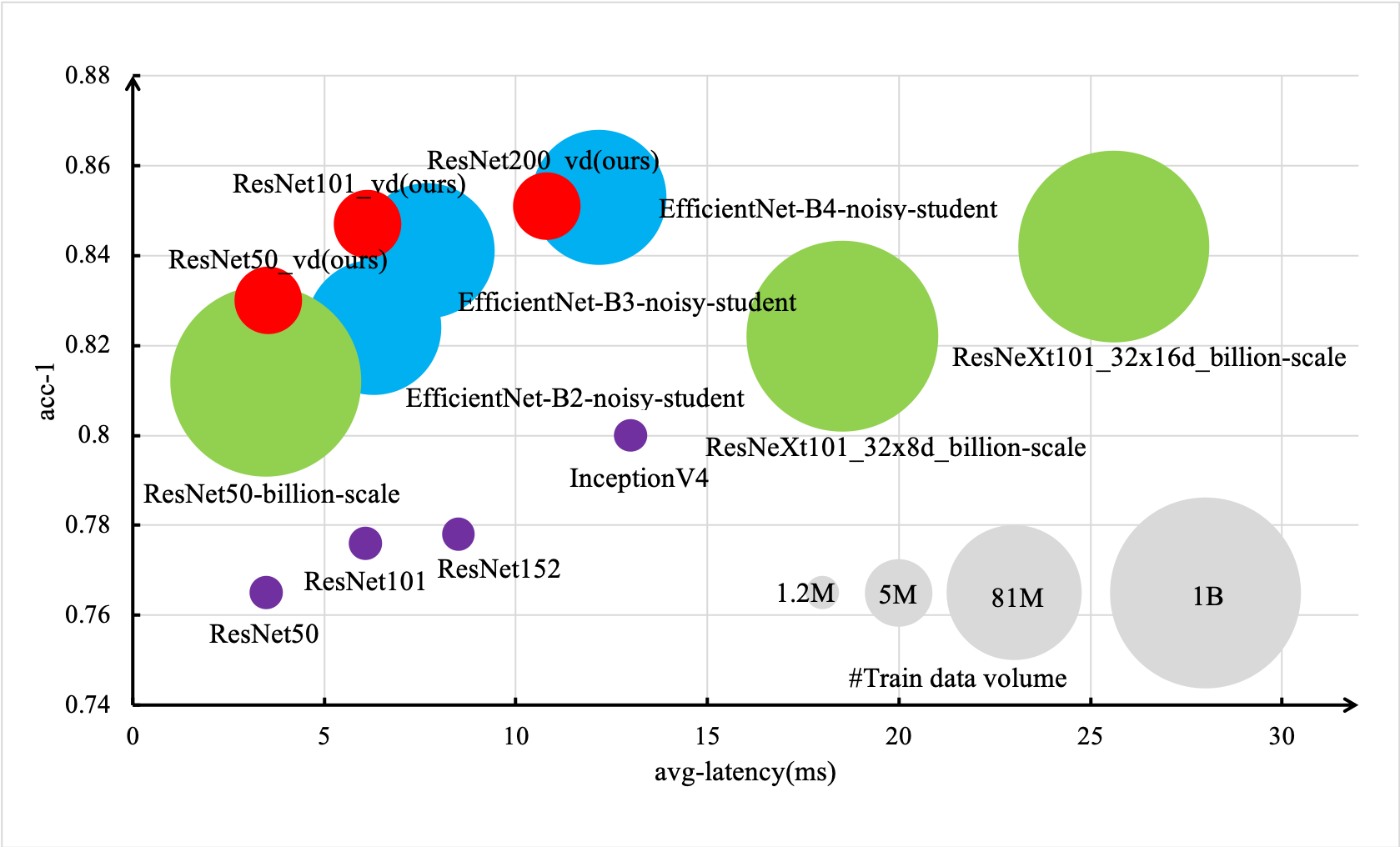}
\caption{ Accuracy-Latency-Unlabeled Data Volume trade-off of different models on ImageNet-1K validation set. In this figure, the average latency is measured on a T4 GPU with a mini-batch of size 1 with TensorRT inference engine at FP32. The size of a circle represents data consumption and the Y-axis value of its center is the top-1 accuracy. The purple, blue, green and red circles denote standard supervised training, Noisy Student \cite{xie2020noisestudent}, Billion-Scale \cite{yalniz2019billion} and our distillation solution, respectively.}
\label{tissert}
\end{figure}

\section{Introduction}

Convolutional Neural Network (CNN) has been developing rapidly in recent years. It has been proven to be the best practice for many computer vision tasks, including image recognition \cite{lin2014mscoco,he2016resnet}, semantic segmentation \cite{chen2017deeplab,deeplabv3}, object detection \cite{liu2016ssd,lin2014mscoco} and so on. The success of CNN majorly comes along with both the prevalence of modern parallel computing devices such as GPUs and the availability of numerous annotated datasets. The former fact makes fast inference of CNNs practical while the later one makes training large CNN models possible. How to further improve performances of backbone CNNs without heavy extra computation cost or without massive manually data annotations then becomes key research topics.  

Toward these topics, there exist multiple solutions. One of the hot directions is to design (or search) better network architectures \cite{he2016resnet,szegedy2016inception,he2019bagtricks,liu2018darts,howard2019mbnetv3}, which can obtain backbones with better efficiency-effectiveness trade-off. The natural question we may raise is how can we further improve the performance of existing off-the-shelf backbones with limited labeling cost, regardless whether they are manually designed with human insights or found by searching from huge network space. Reviewing the literature, a possible answer is to get better pre-trained weights without extra labeling efforts by self-supervision \cite{noroozi2018boostselftraining,he2020contrastivelearning} or semi-supervised training\cite{meanteacher,mixmatch,fixmatch,yalniz2019billion,xie2020noisestudent}. Though effective, these methods have to leverage predefined pretext tasks or huge amount of data to obtain a satisfying pre-trained backbone, and the cost of training phase is very expensive due to the huge training corpus consumption as well as the multiple training stages in need. 
For instances, Noisy Student \cite{xie2020noisestudent} requires up to 81M unique unlabeled images for self-training and it also requires a iterative training process to get more and more robust pseudo labels for unlabeled data; \cite{yalniz2019billion} consumes up to 1 billion images for semi-supervised learning and it also has to perform multiple stages such as teacher-student model pre-training and student model finetuning.

\begin{figure*}[t]
\centering
\includegraphics[width=0.9\linewidth]{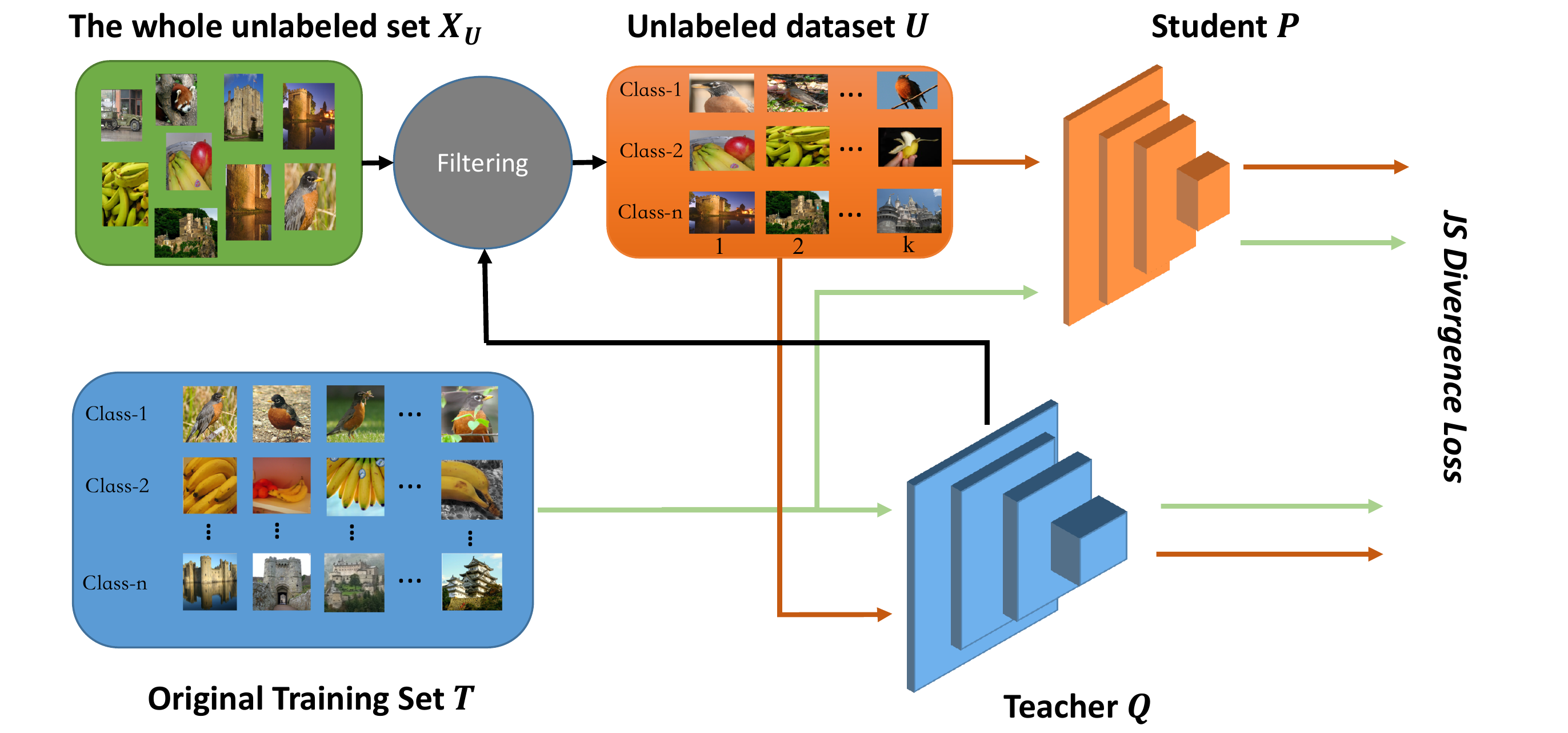}
\caption{Illustration of our overall framework. The knowledge of the teacher model Q is distilled into student P using the training set T and a selected unlabeled data set U. Q is leveraged to filter highly valuable samples from the whole unlabeled set.}
\label{fig-overview}
\end{figure*}

In this paper, we also concentrate our efforts on developing technologies for improving off-the-shelf backbones. The following three questions will be answered in this work. First of all, we challenge whether existing self-supervision solutions are the ultimate choices. Is there any simpler yet effective alternative training paradigm? The second one we would like to study is whether we really need hundreds of millions of unlabeled images to train the backbone. Finally, we will answer whether our proposed alternative, which is much simpler and cheaper, generalizes well to downstream tasks, including transfer learning, object detection and semantic segmentation.

Specifically, we have developed a simple yet effective network distillation framework for this task. Knowledge distillation can directly result in student models for image recognition task. Different from other distillation frameworks \cite{hinton2015distilling,learningfromnoiselabelwithdistillation,yuan2019kd-labelsmooth} which force the student prediction to align with manually labeled one-hot \emph{hard-label} (or its smoothed version), in our framework, student is only required to be consistent with \emph{soft-labels} predicted by teacher model. One of motivations for this design choice is the fact that images can contain multiple objects, thus manually labeled one-hot hard-label cannot well describe such a nature. The other one is that soft-label can get rid of annotations and unlabeled data can be directly utilized for fully exploit the knowledge learned by the teacher model. 
Inspired by theoretical conclusions drawn in the literature, we also find a way to successfully obtain plausible backbone models using much less unlabeled data than existing state-of-the-arts, as can be seen from Figure \ref{tissert}. In more detail, top-1 classification accuracy of 79.0\% and 83\% can be achieved by MobileNetV3-large \cite{howard2019mbnetv3} and ResNet50-D \cite{he2016resnet} with only 4 million extra unlabeled images. We also conduct extensive experiments on downstream tasks, and the results show that the distilled backbones can be leveraged to consistently  improve the performances remarkably on transfer learning, object detection
and semantic segmentation.

In short, we emphasize that this work mainly aims at providing a comprehensive perspective on pushing the limit of existing backbones instead of proposing new ones. The contributions are summarized as follows:
\begin{itemize}
    \item We proposed a simple yet effective distillation framework to improve existing backbones with unlabeled data;
    \item Empirical analysis on how such important factors as teacher model, unlabeled data volume, weight-decay setting \emph{etc.} make a difference are conducted;  
    \item A large variety of models can be significantly improved with only 4M extra unlabeled data and many downstream tasks can remarkably benefit from the resulted backbones.
\end{itemize}


\section{Related Works}

Our work focuses on developing technologies to improve existing backbones, and it is closely related to previous works which make such attempts on this tasks, including network distillation solutions and model pre-training under semi-supervised learning settings or self-supervision settings. 

\textbf{Knowledge Distillation} 
Knowledge distillation has been originally proposed in \cite{hinton2015distilling} to distill knowledge in an ensemble of models into one single model. \cite{romero2014fitnets} used not only the final output but also intermediate hidden layer values of the teacher network to train the student network and showed that using these intermediate layers can improve the performance of deeper and thinner student networks.
In \cite{yim2017fsp} FSP (Flow of Solution Procedure) matrix is proposed to inherit the relationship between features from two layers. There are also other attempts aiming at improving knowledge distillation from perspectives of label-smoothing regularization \cite{yuan2019kd-labelsmooth}, presence of noisy label \cite{xie2020noisestudent} and role-wise data augmentation \cite{fu2020role} and so on. There are a plenty of follow up works on knowledge distillation improvement and its applications. A detailed survey can be found from \cite{gou2020kdsurvey}. The referred knowledge distillation jobs always follow the conventional setting where the prediction of student model should be consistent with the labeled groundtruth. In our work, we simply require the student to be well aligned with the teacher model in terms of final prediction. This enables us to fully distill knowledge of teacher using unlabeled data.

\textbf{Semi-Supervised Learning} 
Supervised learning is verified to be effective, however it is bottlenecked by labor cost of data annotation. Semi-supervised learning (SSL), which can leverage unlabeled data to help model training, has demonstrated great potential. Recently, SSL has been successfully applied in deep learning area. Mean-teacher \cite{meanteacher}, Mix-Match \cite{mixmatch}, UDA \cite{xie2019uda} and Fix-Match \cite{fixmatch} are the recent state-of-the-arts, where unsupervised data augmentation strategies are designed for obtaining regularization from unlabeled data as training signals, thus improving the performance of backbone networks. The methods listed above are usually verified on small datasets such as CIFAR-100 \cite{krizhevsky2009cifar} or small networks. When it comes to big models or large scale datasets, semi-supervised learning is also proved to be effective.In \cite{yalniz2019billion}, by leveraging 1 billion unlabeled images, top-1 accuracy of ResNet-50 on ImageNet benchmark is largely boosted to 81.2\%. This work have concluded with quite many recommended best practices on web-scale semi-supervised learning. In this paper, we further explore how to eliminate the data consumption for saving training costs. Recently, Noisy student \cite{xie2020noisestudent} is proposed to extend self-training and distillation with the use of equal-or-even-larger student model, and an iterative training procedure is designed to obtain state-of-the-art models using 81 million unique unlabeled data. In our work, iterative training is not in need and we can achieve state-of-the-art recognition performance using about 4 million unlabeled data.

\section{Proposed Method}
\label{sec:approach}
\subsection{Overview}
The overall solution is depicted in Figure \ref{fig-overview}. Our distillation framework tries to transfer knowledge learned by a teacher model $Q$ to a student model $P$, with the training dataset $T$ and a carefully collected unlabeled dataset $U$, which is a subset of a much larger unlabeled image gallery $X_U$. In more detail, the teacher model can be any of well trained off-the-shelf ImageNet classification models which are more powerful than the student model. The teacher model is also leveraged to perform unlabeled data filtering in our solution. The filtering step is designed to make sure that 1) valuable unlabeled data is selected to help the distillation process; 2) the selected data is quite different from the validation dataset $V$, ensuring fairness of our comparisons with other solutions as well as the reliability of our experimental results. Both the training set $T$ and the unlabeled set $U$ are used for knowledge distillation. Different from many state-of-the-art solutions where the annotations of $T$ are still used to supervise the predictions of student model, we only force that predictions of the student model and teacher model are well aligned. Such a constraint is achieved by minimizing the Jensen–Shannon divergence. 

\subsection{Problem Formulation}
We denote the training set and validation set of ImageNet-1k \cite{deng2009imagenet} as $T=\{(x_i, y_{x_i})\}$ and $V=\{(x_i, y_{x_i})\}$, respectively, where $x_i$ means a input image and $y_{x_i}$ is the corresponding label. Besides, we have an extra unlabeled dataset $U=\{x_i\}$. 
Our teacher model $Q^*$ is parametered by $\theta_Q^*$, where $\theta_Q^*$ is obtained by:
\begin{equation}
    \theta_Q^* = \min_{\theta_Q} {\Sigma_{(x_i,y_{x_i})\in T}{L\left(y_{x_i}, Q(x_i|\theta_Q)\right)}+\alpha||\theta_Q||_2},
\end{equation}
where $Q(x_i|\theta_Q)$ is the classification probability of $x_i$ generated by teacher model $Q$, $L$ is the cross-entropy loss function and $\alpha$ is the weight decay factor.
Our assumption here is that the teacher model is well trained and its generalization error $R_Q^*$ of is bounded by a small number $R_0$, i.e.,
\begin{equation}
R_Q = \Sigma_{(x_i,y_{x_i})\in V}{L\left(y_{x_i}, Q(x_i|\theta_Q^*)\right)} \leq R_0,
\end{equation}
 
Under our distillation setting, the objective is to obtain an optimal student model $P$ parametered by $\theta_P$ from $Q^*$ by solving the following problem:
\begin{equation}
    \theta_P^* = \min_{\theta_P}{\Sigma_{(x_i)\in T\cup U}{L\left(Q^*(x_i|\theta_Q^*), P(x_i|\theta_P)\right)}+\beta||\theta_P||_2},
\label{eq}
\end{equation}
where $\beta$ is the weigh decay factor for student model $P$.

\subsection{Design Choices}
In this section, we will analysis and explain our design choices in detail, including the selection of teacher model, our choice of using soft-label, our insights in weight decay factor and the way we collecting unlabeled data.
\subsubsection{Teacher Model}
Teacher model plays a central role in our framework. Intuitively, the more powerful the teacher model is, the better the student will be. For a thorough evaluation purpose, we choose to improve a series of backbones (student models) whose target deploy platform ranges from mobile devices to powerful modern GPU devices. In our current implementation, when we aim at improving relatively larger models, the teacher model is ResNeXt101-32x16d \cite{resnext} which is well trained with nearly 1 billion images and its top-1 accuracy (denoted as top-1 acc in the following) is 84.8\% \cite{yalniz2019billion}. When the student model is a lightweight one, such as MobileNet series \cite{howard2017mobilenets,howard2019mbnetv3}, the teacher model we use is a ResNet50-D \cite{he2019bagtricks} improved with our proposed distillation framework and its top-1 acc is as high as 83\%. The reason we use ResNet50-D to improve lightweight models are two-fold: 1) Compared to its students, its accuracy and capacity is sufficiently high,so it is qualified to teach student models; 2) the computation intensity of ResNet50-D is much smaller than other big models, it is very efficient for distillation. 

\subsubsection{Soft-label v.s. Hard-label}
When knowledge distillation was first proposed \cite{hinton2015distilling}, both soft-label generated by teacher model and hard-label annotated with human intelligence were used as supervision information. Follow up works also adopt this paradigm. To improve the performance, many distillation method introduced a hyper-parameter termed as temperature $\tau$ when generating soft-labels. However, if we want to get the best results, the temperature should be well tuned according to different teacher models, respectively. Therefore, obtaining a global optimal $\tau$ is very difficult. 

In this work, we find that very competitive results or even better performance can be achieved using only soft-labels without the parameter $\tau$.
In addition, since the student model only cares about the output of the teacher model in our solution, we can use any data we want to improve performance without introducing any labeling labor. Another motivation of merely using the soft-labels is inspired by the properties of natural images. Specifically, natural images can have multiple tags, for example, a cow on the grass can be tagged as ``cow'' or ``grass'', most of popular datasets provides only a single label for each image and such hard-label is not sufficient to describe the information of the image. Meanwhile, soft-label predicted by a well-trained model can generate a probability distribution over the tag system. Entropy of soft-labels can be larger than hard-labels, meaning that more information can be provided. 


\subsubsection{Weight-Decay}
Regularization in the optimization of deep neural networks is often critical to avoid undesirable over-fitting. One of the most popular regularization algorithms is to impose L2 penalty on the model parameters as shown in Eq.\ref{eq}, called weight- decay.

Appropriately reducing the value of $\beta$ can result in smaller training loss. Potentially, the validation accuracy can be improved when the training loss is reduced without heavy over-fitting. Intuitively, because in our framework the student model is required to fit a much more complex mapping, where image to soft-label which has much larger entropy than hard-label, reducing weight decay to some extent would not result in sever over-fitting. 

According to \cite{golowich2018size}, the model generalization error scales (with high probability) as:
\begin{equation}
    \mathcal{O}\left(\frac{B2^d\Pi_{j=1}^dM_F(j)}{\sqrt{m}}\right),
\end{equation}
where $d$ is network depth, $m$ is number of training samples, $M_F(j)$ is Frobenius norms of weight parameter matrix at the $j^{th}$ layer and $B$ is a constant. Recall that our framework can get rid of annotations, so we can have sufficiently large $m$. From this formula, we can see that, our solution can also have very small over-fitting risk due to the increasing of $m$ can be as large as we want, although reducing weight decay can result in increased $M_F(j)$'s.
In the ablation experiment, we also further illustrate the importance of reducing $\beta$ in knowledge distillation through experiments.

\subsubsection{Unlabeled Data Collection}
For labeled data, we use the full training data of ImageNet-1K, which has a total of 1.2M images. In order to improve performance, we need to collect an unlabeled dataset in our framework. In detail, the unlabeled data gallery $X_U$ we use is the ImageNet-22K, from which 4M images are selected to form $U$ with the help of the teacher model of ResNeXt101-32x16d \cite{yalniz2019billion}. 

The filtering procedure is performed as follows:
First of all, We filter out images that are visually similar with validation images of ImageNet-1K from the ImageNet-22K. In this step, the well-known SIFT \cite{sift} is used to match similar image pairs from the two image sets. The purpose of this step is to prevent training the student model on images similar to the validation set and keep the evaluation results fair enough. Then, we use the teacher model to obtain prediction results for the images of ImageNet-22k. Finally, we sort the images within each category according to their score and take the top k images of each category to form the final unlabeled data $U$. Currently, k is empirically set to 4000, so our unlabeled data has a total of 4M images and in total 5.2M images are used for distillation.

\section{Experiment}

\subsection{Datasets and Evaluation Metrics}
In this paper, we use ImageNet 2012 ILSVRC dataset \cite{deng2009imagenet} to evaluate our method and compare with other state-of-the-arts. The dataset consists of 1000 categories and 1.28 million training images in total, which is considered to be the most convincing one in the field of image classification. It is also the most preferred dataset to pre-train models for downstream tasks in the computer vision community. 
Analogy to most image classification scenarios, we report top-1 accuracy on the validation partition as the evaluation metric.

\subsection{Implementation Details}
When training on ImageNet, we follow standard practice and perform data augmentation with random-size cropping to 224$\times$224 pixels and random horizontal flipping. Optimization is performed using SGD with momentum 0.9 and a mini-batch size of 256. Weight decay is empirically set to 4e-5 for large student models and 1e-5 for small student models and initial learning rate is 0.1. As commonly known that bigger models need more regularization, we also use  AutoAugment \cite{cubuk2019autoaugment} on large student models. For declining learning rate, we use a cosine learning rate schedule with the first 5 epochs reserved for warm-up. The training epoch of distillation training with labeled and unlabeled data in the first stage is 360, and for further improvement, we finetune the student with only labeled data for extra 30 epochs.

\subsection{Comparison with State-of-the-Arts}
We first compare our proposed method with state-of-the-arts \cite{yalniz2019billion,xie2020noisestudent} on the ImageNet dataset as commonly done in previous works.
As shown in Table \ref{table:one}, our algorithm performs consistently better than other variants of ResNet50 backbone, such as using auto-augment strategies including cutout \cite{cubuk2019autoaugment}, mixup \cite{mixmatch}, etc,.
Specifically, Without finetuning, our method improves ResNet50-D from 79\% to 82.2\%. Finetuing brings an extra gain of 0.8\%. When ResFix\cite{touvron2019fixingresolution} which is proposed to fix the training-test resolution discrepancy 
is further adopted, top-1 acc will be boosted up to 84\%, which is significantly higher than 79.1\% reported in \cite{he2019bagtricks}. Note that the remarkable improvement comes at the cost of only 4M extra unlabeled data, consuming 5.2M images in total.

Moreover, our method outperforms Billion-scale \cite{yalniz2019billion} by an absolute gain of 1.8\%, which demands huge training cost for semi-supervised learning 940M of unlabeled images on the training phase. As for Noisy Student \cite{xie2020noisestudent}, it uses a unlabeled dataset 81M images for iteratively training. Considering that the inference latency is comparable with ResNet50-D on a T4 GPU, here we report the performance of EfficientNet-B0 for comparison and top-1 acc of Noisy Student is 78.5\%. To illustrate the strong performance of our method, we also compared the performance of EfficientNet-B1, whose latency is 1.5 times that of ResNet50-D.
On the contrary, our solution need requires 4M extra unlabeled data to exploit the knowledge learned by the teacher model, which is the overall distillation is much computational efficient.

\begin{table}[h]
\begin{center}
\begin{tabular}{p{140pt}|p{40pt}|p{25pt}}
\hline
Method & Training Data Volume & Top-1 Acc(\%) \\
\hline\hline
ResNet50 & 1.2M & 76.5 \\
+cutout & 1.2M & 77.1 \\
+mixup & 1.2M & 77.4 \\
+cutmix & 1.2M & 78.6 \\
+gridmask  & 1.2M & 78.7 \\
+cutmix+label-smooth & 1.2M & 79.0 \\
ResNet50-D+mixup+label-smooth & 1.2M & 79.1 \\
\hline
ResNet50(Billion-scale) & 940M & 81.2 \\
EfficientNet-B0(Noisy-Student) &81M &78.8 \\
EfficientNet-B1(Noisy-Student) &81M &81.5 \\
\hline
Ours(ResNet50-D) w/o finetune & \textbf{5.2M} & \textbf{82.2} \\
Ours(ResNet50-D) w finetune & \textbf{5.2M} & \textbf{83.0} \\
Ours(ResNet50-D+ResFix) & \textbf{5.2M} & \textbf{84.0} \\
\hline
\end{tabular}
\end{center}
\caption{Comparison of top-1 acc with state of the art methods.}
\label{table:one}
\end{table}

\subsection{Performance of different models}
In order to provide a more comprehensive understanding, we provide results of different backbones, including models designed for mobile devices and GPU servers, improved using our proposed distillation solution. The experimental results are summarized in Table \ref{table:two}. In this table, the baseline top-1 acc reported is achieved by following the standard ImageNet-1K training procedure. It is clear that our solution can consistently increase model accuracy by a very large margin, no matter it is a lightweight mobile model or it is a very powerful model as big as ResNet200-D. Especially, our method can boost the performance of MobileNetV3-large from 75.3\% to 79\% and such a performance is comparable with the ResNet50-D baseline, meaning that using our distillation framework, equivalent accuracy can be achieved with much cheap models. Meanwhile, ResNet200-
D is improved by an absolute gain of 4.1\%, it further validates that our solution is robust and will not saturate when the student model gets larger. 

\begin{table}[t]
\begin{center}
\begin{tabular}{p{90pt}p{40pt}p{60pt}}
\hline
Model & Top-1 Acc(\%) & Top-1 Acc(\%)(Ours)  \\
\hline\hline
MobileNetV1 & 70.9 & \textbf{77.9 \textcolor[RGB]{0,128,0}{(+7.0)}}  \\
MobileNetV2 & 72.1 & \textbf{76.4 \textcolor[RGB]{0,128,0}{(+4.3)}}\\
MobileNetV3-small & 68.2 & \textbf{71.3 \textcolor[RGB]{0,128,0}{(+3.1)}}\\
MobileNetV3-large & 75.3 & \textbf{79.0 \textcolor[RGB]{0,128,0}{(+3.7)}} \\
HRNet-18W-C & 76.9 & \textbf{81.2 \textcolor[RGB]{0,128,0}{(+4.3)}}\\
SE-HRNet-64W-C & 80.6 & \textbf{84.8 \textcolor[RGB]{0,128,0}{(+4.2)}}\\
ResNet50-D & 79.1 & \textbf{83.0 \textcolor[RGB]{0,128,0}{(+3.9)}}\\
ResNet101-D & 80.3 & \textbf{83.7 \textcolor[RGB]{0,128,0}{(+3.4)}}\\
ResNet200-D & 80.9 & \textbf{85.0 \textcolor[RGB]{0,128,0}{(+4.1)}}\\

\hline
\end{tabular}
\end{center}
\caption{The performance of different models, our distillation scheme will greatly improve the accuracy of different models, each individual backbone has achieved its new state-of-the-art performance.}
\label{table:two}
\end{table}

\subsection{Ablation Study} 
Here we study what factors will influence the distillation performance in our framework and how they make a difference. Experiments are conducted using the mobile settings, \emph{i.e.}, the teacher is ResNet50-D and the student is MobileNetV3, because larger models are too time-consuming to complete all the below experiments.

\textbf{Impact of different teacher models}
We evaluate the performance of different teacher models using MobileNetV3-large as the student. Intuitively, a better teacher model tends to guide a student. This intuition does not heavily conflict with our experiment result shown in Table \ref{table:3}. However, when the accuracy of the teacher model exceeds a certain range, the student model will not be further improved. From the table, we can see that MobileNetV3-large can achieve top-1 accuracy of 76.8\% and 78.4\% when using ResNet50-D as teacher whose top-1 acc of 79.1\% and 83\%, respectively. When the accuracy of the teacher model is further improved, the accuracy of MobileNetV3-large will no longer increase. This result also makes sense, because the capacity of student model is limited, its performance upper bound is also limited. Such experimental result also suggest an appropriate teacher model is important to our framework in terms of accuracy-efficiency trade-off, especially when computational resources are a central concern.

\begin{table}[t]
\begin{center}
\begin{tabular}{p{80pt}p{50pt}p{50pt}}
\hline
Teacher Model & Teacher Model Top-1 Acc(\%) & Student Model Top-1 Acc(\%) \\
\hline\hline
ResNet50-D & 79.1 & 76.8 \\
ResNet50-D (ours) & \textbf{83.0} & \textbf{78.4} \\
ResNet101-D & 83.7 & 78.4 \\
ResNet200-D & 85.0 & 78.4 \\
\hline
\end{tabular}
\end{center}
\caption{The accuracy of MobileNetV3-large under different teacher models, the number of distilled training epoch is 120, weight-decay is 1e-5.}
\label{table:3}
\end{table}

\textbf{Convergence}
We study the convergence of our framework by evaluating top-1 acc of student network at different training iterations. In this experiment, the teacher model we select is ResNet50-D with top-1 accuracy of 83\%, while the student network is MobileNetV3-large.
We fix the weight decay of 1e-5.
Generally, the observation is, the more epochs the better.
As seen in Table \ref{table:4}, when the number of training epoch increases from 50 to 200, the top-1 accuracy improves gradually meanwhile when it reaches 360, the performance improvement becomes very marginal especially when an extra 30 epochs of finetuning is carried out on ImageNet-1K, showing that 360 epochs can ensure the knowledge of teacher is sufficiently exploited.

\begin{table}[h]
\begin{center}
\begin{tabular}{p{50pt}p{50pt}p{60pt}}
\hline
epochs & Top-1 Acc(\%) & Top-1 Acc(\%)(finetune)  \\
\hline\hline
50 & 76.98 & 77.58 \\
100 & 77.77 & 78.29 \\
200 & 78.29 & 78.75 \\
300 & 78.42 & 78.90 \\
360 & \textbf{78.54} & \textbf{79.00} \\
400 & 78.56 & 79.01 \\
\hline
\end{tabular}
\end{center}
\caption{
Top-1 Acc of student network at different epochs.}
\label{table:4}
\end{table}

\begin{figure}[h]
\centering
\includegraphics[width=0.9\columnwidth]{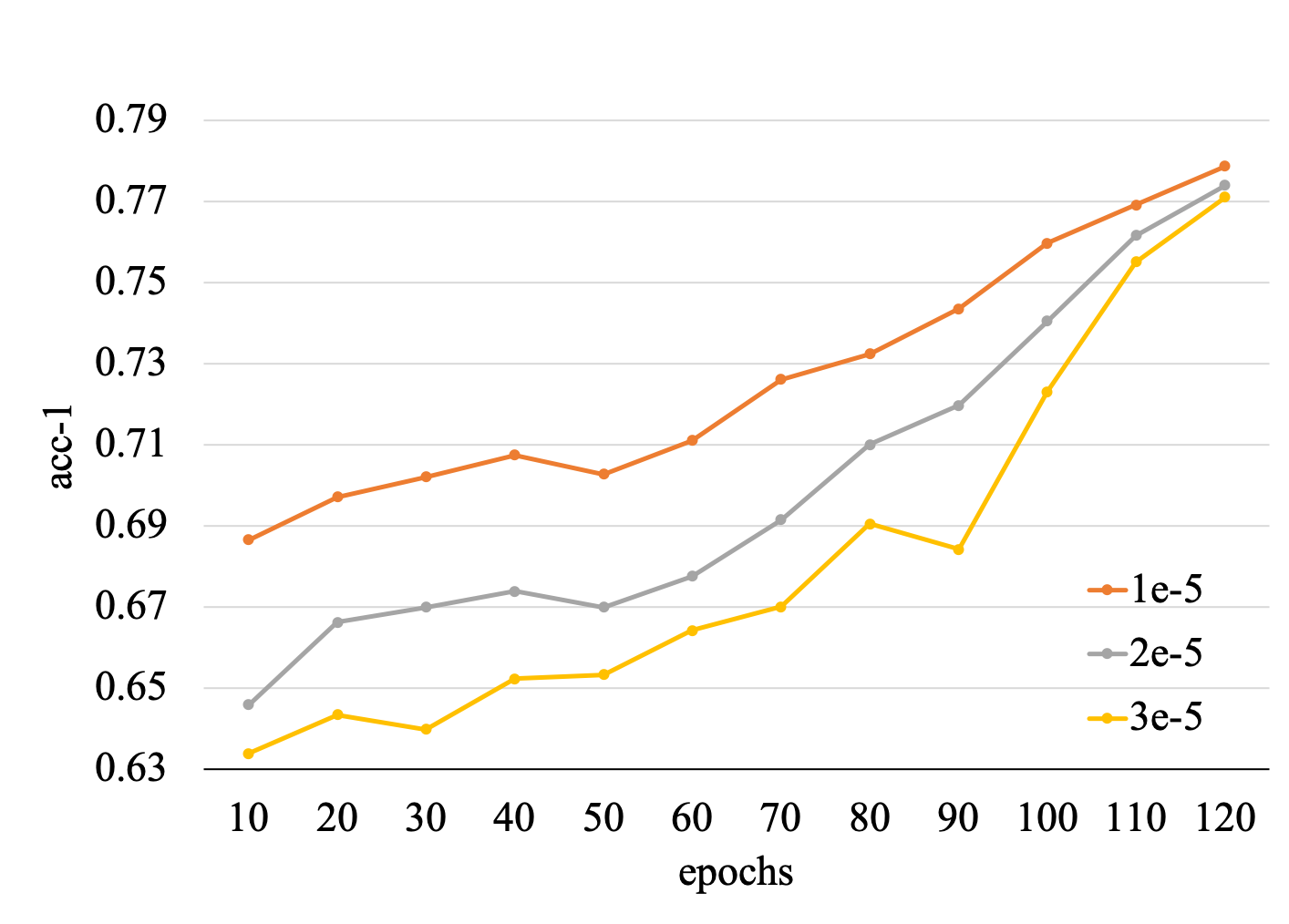}
\caption{Convergence curves of different distillation experiments with various weight decay factors.}
\label{curve}
\end{figure}

\textbf{Impact of different weight decay}
Weight decay play a key role to regularize models and prevent over-fitting. In this experiment, we change the weight decay factor $\beta$ in our framework where the teacher is ResNet50-D with top-1 acc of 83\% and the student network is MobileNetV3-large. The final results of our model (with or without finetuing on ImageNet-1k for 30 epochs) is shown in Table \ref{table:5} and the convergence behavior is depicted in Figure \ref{curve}. The results show that decreasing $\beta$ from 3e-5 to 1e-5 can speed up the model convergence and the top-1 accuracy will also be increased without over-fitting risk. Empirically, 1e-5 is a good setting for $\beta$ when distilling knowledge of ResNet50-D (ours) into MobileNetV3-large, the weight decay factor of 4e-5 is also empirically tuned and determined when distilling ResNeXt101-32x16d into larger student models such as ResNet50-D, HRNet-18W-C \cite{hrnet} and HRNet with squeeze-and-excitation \cite{senet} \emph{etc.}

\begin{table}[h]
\begin{center}
\begin{tabular}{p{30pt}p{40pt}p{60pt}}
\hline
weight-decay & Top-1 Acc(\%) & Top-1 Acc(\%)(finetune)  \\
\hline\hline
3e-5 & 77.11 & 77.63 \\
2e-5 & 77.40 & 77.95 \\
1e-5 & \textbf{77.88} & \textbf{78.36} \\
\hline
\end{tabular}
\end{center}
\caption{
Top-1 Acc of student network under different weight-decay. The number of distilled training epochs is 120.}
\label{table:5}
\end{table}

\textbf{Impact of different unlabeled data volume}
In this experiment, we validate that 4M data is already sufficient to improve backbone via our distillation framework. As can be seen from Table \ref{table:6}, when the unlabeled data is increased from 2M to 4M the accuracy of MobileNetV3-large gains slowly. However, when using more data, the training time and cost will increase linearly, we do not using more data in our current implementation for well balancing the performance and cost.

\begin{table}[h]
\begin{center}
\begin{tabular}{p{60pt}p{50pt}p{60pt}}
\hline
Unlabeled Data Volume  &Top-1 Acc(\%) &Top-1 Acc(\%)(finetune)  \\
\hline\hline
2M & 77.71 & 78.06 \\
3M & 77.80 & 78.29\\
4M & 77.88 & \textbf{78.36} \\
5M & 77.92 & 78.38\\
\hline
\end{tabular}
\end{center}
\caption{The accuracy of MobileNetV3-large under different Unlabeled data volumes. Among them, the Teacher Model is ResNet50-D with top-1 acc of 83\%, and the labeled data of each experiment is 1.2M (ImageNet-1k), the number of distilled training epochs is 120.}
\label{table:6}
\end{table}

\subsection{Improvements in Downstream Tasks}
We also conduct experiments on several downstream tasks, including image classification, object detection and semantic segmentation, to verify the effectiveness of our distilled model as initialization.

\textbf{Classification}
We first investigate generalization of the proposed distilled model and performs transfer learning experiments on several image classification datasets, including FGVC-Aircraft-2013b ~\cite{maji2013fine}, CIFAR-100 ~\cite{krizhevsky2009learning}, DTD ~\cite{cimpoi2014describing} and SUN397 ~\cite{xiao2010sun}.

We finetune the entire network using the weights of the pre-trained network as
initialization.
For each dataset, we search the best hyper-parameters with ResNet50-D baseline on validation set to ensure fair comparison.
Note that for all experiments, after searching, we use the same hyper-parameters and strategies for training, except for changing the pre-trained model.

Table \ref{table:7} summaries the comparison results with baseline ResNet50-D model in terms of top-1 classification accuracy.
As can be seen, our method beats ResNet50-D baseline on all datasets with considerable margins, demonstrating that our distilled model can be well generalized in these downstream tasks.
An interesting finding is that, the lower the baseline accuracy is, the larger our model improves.
For example, for the SUN397 dataset, we outperform baseline by a margin of 4.34\%.
While for FGVC-Aircraft-2013b, the baseline accuracy is high enough (88.98\%),  our method achieves an improvement of 1.02\%.

\begin{table}[h]
\begin{center}
\begin{tabular}{p{60pt}p{60pt}p{80pt}}
\hline
Dataset & Baseline Top-1 Acc(\%) & Ours Top-1 Acc(\%) \\
\hline\hline
FGVC-Aircraft-2013b  & 88.98 & \textbf{90.00 \textcolor[RGB]{0,128,0}{(+1.02)}} \\
CIFAR-100 & 86.50 & \textbf{87.58 \textcolor[RGB]{0,128,0}{(+1.04)}} \\
DTD & 76.48 & \textbf{77.71 \textcolor[RGB]{0,128,0}{(+1.23)}} \\
SUN397 & 64.02 & \textbf{68.36 \textcolor[RGB]{0,128,0}{(+4.34)}}  \\
\hline
\end{tabular}
\end{center}
\caption{Comparison of the acc1 of the ordinary ResNet50-D pre-trained model and the distilled version on different datasets, except for replacing the pre-trained model, we use the same training strategy.}
\label{table:7}
\end{table}

\textbf{Object Detection}
In this experiment, we compare our distilled models with baseline ResNet50-D on object detection tasks based on PaddleDetection$\footnote{https://github.com/PaddlePaddle/PaddleDetection}$.
All experiments are conducted on the commonly used MS-COCO2017 \cite{lin2014microsoft} dataset, which involves 80 object categories.
Following the standard COCO metric, we report the average mAP at IoU of 0.5:0.05:0.95.
The official train/val split is used for training and validation.
As for the detection framework, we select two-stage, one-stage and anchor-free methods, i.e., Faster-RCNN-FPN \cite{ren2015faster}, YOLOv3 \cite{redmon2018yolov3} and TTFNet \cite{liu2020training}, to prove the widely effectiveness of our distilled backbone.


Table \ref{table:8} demonstrates the experiment results on COCO minival2017, results are obtained by single-scale training/testing mechanism. With our improved backbone, object detection performance can be consistently increased, regardless of what kind of detector is used. Our backbone can increase mAP@[0.5,0.95] of Faster-RCNN-FPN, YoLOv3, TTFNet and RetinaNet by 1.5\%, 1.1\%, 2.1\% and 1.1\%, respectively. Besides, our distillation solution is also complementary for deformable convolutional, adding deformable convolutional v2 (DCNv2)\cite{zhu2019deformable} on the basis of ResNet50-D(ours)-YOLOv3 can still gain by 1.2\%.
This suggests that pre-trained backbone obtained from our distillation solution can well improve object detection tasks.

\begin{table}[h]
\begin{center}
\begin{tabular}{p{80pt}p{50pt}p{30pt}p{45pt}}
\hline
Model & Backbone &Baseline mAP(\%) & Ours mAP(\%) \\
\hline\hline
Faster-RCNN-FPN & ResNet50-D & 34.8 & \textbf{36.3 \textcolor[RGB]{0,128,0}{(+1.5)}} \\
RetinaNet & ResNet50-D & 37.0 & \textbf{38.1 \textcolor[RGB]{0,128,0}{(+1.1)}} \\
TTFNet & ResNet50-D & 33.2 & \textbf{35.3 \textcolor[RGB]{0,128,0}{(+2.1)}} \\
YOLOv3 & ResNet50-D & 37.4 & \textbf{38.5 \textcolor[RGB]{0,128,0}{(+1.1)}} \\
YOLOv3-DCNv2 & ResNet50-D & 39.1 & \textbf{40.3 \textcolor[RGB]{0,128,0}{(+1.2)}} \\
\hline

\end{tabular}
\end{center}
\caption{Comparison of the detection mAP on MS COCO dataset between ordinary ResNet50-D pretrained model and its distilled version with our solution.} 
\label{table:8}
\end{table}

\textbf{Semantic Segmentation}
We also report the semantic segmentation results on Cityscapes dataset\cite{cordts2016cityscapes}. It contains 5000 high-quality labeled images, among which 2,975 images consists of the training set and 500 images form the validation set.

We implemented it based on PaddleSeg$\footnote{https://github.com/PaddlePaddle/PaddleSeg}$.We follow the same training protocol of \cite{zhao2017pyramid} and following prior work\cite{chen2018encoder,yuan2019object} to report mIOU evaluated on the validation set for comparison. Table \ref{table:9} provides the detailed results of different  semantic  segmentation heads with different pretrain models. 
With different semantic segmentation heads, our improved backbone models has consistently shown great advantages compared to their baseline counterparts. Even on a state-of-the-art head like OCRNet, we can still have an obvious improvement.

\begin{table}[h]
\begin{center}
\begin{tabular}{p{50pt}p{65pt}p{30pt}p{50pt}}
\hline
Model & Backbone & Baseline mIoU(\%) &  Ours mIoU(\%) \\
\hline\hline
FCN & HRNetV2-W18 & 78.83 & \textbf{80.38\textcolor[RGB]{0,128,0}{(+1.55)}}\\
FCN & ResNet50-D & 74.87 & \textbf{77.03\textcolor[RGB]{0,128,0}{(+2.16)}}\\
Deeplabv3+ & ResNet50-D & 78.06 & \textbf{79.51\textcolor[RGB]{0,128,0}{(+1.45)}}\\
OCRNet & HRNetV2-W18 & 80.12 & \textbf{80.44\textcolor[RGB]{0,128,0}{(+0.32)}} \\

\hline
\end{tabular}
\end{center}
\caption{mIoU values of different semantic segmentation heads using different pre-train models.}

\label{table:9}
\end{table}

\section{Conclusion and Discussion}
In this paper, we provide positive answer about whether current self-supervision solutions can be alternated by more simple and effective solution. To offer a possible solution, we develop a distillation based framework to improve existing backbones. Our proposed distillation solution only requires prediction of student model is well aligned with that of the teacher model, thus unlabeled data can be easily leveraged to support fully exploitation of the knowledge learned by the teacher. In this paper, we also answer that it is possible to use only 4M unlabeled data to get backbone models significantly improved. Extensive experiments are conducted to offer empirically guidance on our design choices and effectiveness are validated on multiple computer vision tasks.

\textbf{Acknowledgement.}
This work was supported by the National Key Research and Development Project of China (2020AAA0103500).


{\small
\bibliographystyle{main}
\bibliography{egbib}

\begin{thebibliography}{10}\itemsep=-1pt

\bibitem{mixmatch}
David Berthelot, Nicholas Carlini, Ian Goodfellow, Nicolas Papernot, Avital
  Oliver, and Colin~A Raffel.
\newblock Mixmatch: A holistic approach to semi-supervised learning.
\newblock In {\em Advances in Neural Information Processing Systems}, pages
  5049--5059, 2019.

\bibitem{chen2017deeplab}
Liang-Chieh Chen, George Papandreou, Iasonas Kokkinos, Kevin Murphy, and Alan~L
  Yuille.
\newblock Deeplab: Semantic image segmentation with deep convolutional nets,
  atrous convolution, and fully connected crfs.
\newblock {\em IEEE transactions on pattern analysis and machine intelligence},
  40(4):834--848, 2017.

\bibitem{deeplabv3}
Liang-Chieh Chen, George Papandreou, Florian Schroff, and Hartwig Adam.
\newblock Rethinking atrous convolution for semantic image segmentation.
\newblock {\em arXiv preprint arXiv:1706.05587}, 2017.

\bibitem{chen2018encoder}
Liang-Chieh Chen, Yukun Zhu, George Papandreou, Florian Schroff, and Hartwig
  Adam.
\newblock Encoder-decoder with atrous separable convolution for semantic image
  segmentation.
\newblock In {\em Proceedings of the European conference on computer vision
  (ECCV)}, pages 801--818, 2018.

\bibitem{cimpoi2014describing}
Mircea Cimpoi, Subhransu Maji, Iasonas Kokkinos, Sammy Mohamed, and Andrea
  Vedaldi.
\newblock Describing textures in the wild.
\newblock In {\em Proceedings of the IEEE Conference on Computer Vision and
  Pattern Recognition}, pages 3606--3613, 2014.

\bibitem{cordts2016cityscapes}
Marius Cordts, Mohamed Omran, Sebastian Ramos, Timo Rehfeld, Markus Enzweiler,
  Rodrigo Benenson, Uwe Franke, Stefan Roth, and Bernt Schiele.
\newblock The cityscapes dataset for semantic urban scene understanding.
\newblock In {\em Proceedings of the IEEE conference on computer vision and
  pattern recognition}, pages 3213--3223, 2016.

\bibitem{cubuk2019autoaugment}
Ekin~D Cubuk, Barret Zoph, Dandelion Mane, Vijay Vasudevan, and Quoc~V Le.
\newblock Autoaugment: Learning augmentation strategies from data.
\newblock In {\em Proceedings of the IEEE conference on computer vision and
  pattern recognition}, pages 113--123, 2019.

\bibitem{deng2009imagenet}
Jia Deng, Wei Dong, Richard Socher, Li-Jia Li, Kai Li, and Li Fei-Fei.
\newblock Imagenet: A large-scale hierarchical image database.
\newblock In {\em 2009 IEEE conference on computer vision and pattern
  recognition}, pages 248--255. Ieee, 2009.

\bibitem{fu2020role}
Jie Fu, Xue Geng, Zhijian Duan, Bohan Zhuang, Xingdi Yuan, Adam Trischler, Jie
  Lin, Chris Pal, and Hao Dong.
\newblock Role-wise data augmentation for knowledge distillation.
\newblock {\em arXiv preprint arXiv:2004.08861}, 2020.

\bibitem{golowich2018size}
Noah Golowich, Alexander Rakhlin, and Ohad Shamir.
\newblock Size-independent sample complexity of neural networks.
\newblock In {\em Conference On Learning Theory}, pages 297--299. PMLR, 2018.

\bibitem{gou2020kdsurvey}
Jianping Gou, Baosheng Yu, Stephen~John Maybank, and Dacheng Tao.
\newblock Knowledge distillation: A survey.
\newblock {\em arXiv preprint arXiv:2006.05525}, 2020.

\bibitem{he2020contrastivelearning}
Kaiming He, Haoqi Fan, Yuxin Wu, Saining Xie, and Ross Girshick.
\newblock Momentum contrast for unsupervised visual representation learning.
\newblock In {\em Proceedings of the IEEE/CVF Conference on Computer Vision and
  Pattern Recognition}, pages 9729--9738, 2020.

\bibitem{he2016resnet}
Kaiming He, Xiangyu Zhang, Shaoqing Ren, and Jian Sun.
\newblock Deep residual learning for image recognition.
\newblock In {\em Proceedings of the IEEE conference on computer vision and
  pattern recognition}, pages 770--778, 2016.

\bibitem{he2019bagtricks}
Tong He, Zhi Zhang, Hang Zhang, Zhongyue Zhang, Junyuan Xie, and Mu Li.
\newblock Bag of tricks for image classification with convolutional neural
  networks.
\newblock In {\em Proceedings of the IEEE Conference on Computer Vision and
  Pattern Recognition}, pages 558--567, 2019.

\bibitem{hinton2015distilling}
Geoffrey Hinton, Oriol Vinyals, and Jeff Dean.
\newblock Distilling the knowledge in a neural network.
\newblock {\em arXiv preprint arXiv:1503.02531}, 2015.

\bibitem{howard2019mbnetv3}
Andrew Howard, Mark Sandler, Grace Chu, Liang-Chieh Chen, Bo Chen, Mingxing
  Tan, Weijun Wang, Yukun Zhu, Ruoming Pang, Vijay Vasudevan, et~al.
\newblock Searching for mobilenetv3.
\newblock In {\em Proceedings of the IEEE International Conference on Computer
  Vision}, pages 1314--1324, 2019.

\bibitem{howard2017mobilenets}
Andrew~G Howard, Menglong Zhu, Bo Chen, Dmitry Kalenichenko, Weijun Wang,
  Tobias Weyand, Marco Andreetto, and Hartwig Adam.
\newblock Mobilenets: Efficient convolutional neural networks for mobile vision
  applications.
\newblock {\em arXiv preprint arXiv:1704.04861}, 2017.

\bibitem{senet}
Jie Hu, Li Shen, and Gang Sun.
\newblock Squeeze-and-excitation networks.
\newblock In {\em Proceedings of the IEEE conference on computer vision and
  pattern recognition}, pages 7132--7141, 2018.

\bibitem{krizhevsky2009learning}
Alex Krizhevsky, Geoffrey Hinton, et~al.
\newblock Learning multiple layers of features from tiny images.
\newblock 2009.

\bibitem{krizhevsky2009cifar}
A Krizhevsky, V Nair, and G Hinton.
\newblock Cifar-100 dataset (canadian institute for advanced research), 2009.

\bibitem{learningfromnoiselabelwithdistillation}
Yuncheng Li, Jianchao Yang, Yale Song, Liangliang Cao, Jiebo Luo, and Li-Jia
  Li.
\newblock Learning from noisy labels with distillation.
\newblock In {\em Proceedings of the IEEE International Conference on Computer
  Vision}, pages 1910--1918, 2017.

\bibitem{lin2014mscoco}
Tsung-Yi Lin, Michael Maire, Serge Belongie, James Hays, Pietro Perona, Deva
  Ramanan, Piotr Doll{\'a}r, and C~Lawrence Zitnick.
\newblock Microsoft coco: Common objects in context.
\newblock In {\em European conference on computer vision}, pages 740--755.
  Springer, 2014.

\bibitem{lin2014microsoft}
Tsung-Yi Lin, Michael Maire, Serge Belongie, James Hays, Pietro Perona, Deva
  Ramanan, Piotr Doll{\'a}r, and C~Lawrence Zitnick.
\newblock Microsoft coco: Common objects in context.
\newblock In {\em European conference on computer vision}, pages 740--755.
  Springer, 2014.

\bibitem{liu2018darts}
Hanxiao Liu, Karen Simonyan, and Yiming Yang.
\newblock Darts: Differentiable architecture search.
\newblock {\em arXiv preprint arXiv:1806.09055}, 2018.

\bibitem{liu2016ssd}
Wei Liu, Dragomir Anguelov, Dumitru Erhan, Christian Szegedy, Scott Reed,
  Cheng-Yang Fu, and Alexander~C Berg.
\newblock Ssd: Single shot multibox detector.
\newblock In {\em European conference on computer vision}, pages 21--37.
  Springer, 2016.

\bibitem{liu2020training}
Zili Liu, Tu Zheng, Guodong Xu, Zheng Yang, Haifeng Liu, and Deng Cai.
\newblock Training-time-friendly network for real-time object detection.
\newblock In {\em AAAI}, pages 11685--11692, 2020.

\bibitem{sift}
David~G Lowe.
\newblock Object recognition from local scale-invariant features.
\newblock In {\em Proceedings of the seventh IEEE international conference on
  computer vision}, volume~2, pages 1150--1157. Ieee, 1999.

\bibitem{maji2013fine}
Subhransu Maji, Esa Rahtu, Juho Kannala, Matthew Blaschko, and Andrea Vedaldi.
\newblock Fine-grained visual classification of aircraft.
\newblock {\em arXiv preprint arXiv:1306.5151}, 2013.

\bibitem{noroozi2018boostselftraining}
Mehdi Noroozi, Ananth Vinjimoor, Paolo Favaro, and Hamed Pirsiavash.
\newblock Boosting self-supervised learning via knowledge transfer.
\newblock In {\em Proceedings of the IEEE Conference on Computer Vision and
  Pattern Recognition}, pages 9359--9367, 2018.

\bibitem{redmon2018yolov3}
Joseph Redmon and Ali Farhadi.
\newblock Yolov3: An incremental improvement.
\newblock {\em arXiv preprint arXiv:1804.02767}, 2018.

\bibitem{ren2015faster}
Shaoqing Ren, Kaiming He, Ross Girshick, and Jian Sun.
\newblock Faster r-cnn: Towards real-time object detection with region proposal
  networks.
\newblock In {\em Advances in neural information processing systems}, pages
  91--99, 2015.

\bibitem{romero2014fitnets}
Adriana Romero, Nicolas Ballas, Samira~Ebrahimi Kahou, Antoine Chassang, Carlo
  Gatta, and Yoshua Bengio.
\newblock Fitnets: Hints for thin deep nets.
\newblock {\em arXiv preprint arXiv:1412.6550}, 2014.

\bibitem{fixmatch}
Kihyuk Sohn, David Berthelot, Chun-Liang Li, Zizhao Zhang, Nicholas Carlini,
  Ekin~D Cubuk, Alex Kurakin, Han Zhang, and Colin Raffel.
\newblock Fixmatch: Simplifying semi-supervised learning with consistency and
  confidence.
\newblock {\em arXiv preprint arXiv:2001.07685}, 2020.

\bibitem{szegedy2016inception}
Christian Szegedy, Sergey Ioffe, Vincent Vanhoucke, and Alex Alemi.
\newblock Inception-v4, inception-resnet and the impact of residual connections
  on learning.
\newblock {\em arXiv preprint arXiv:1602.07261}, 2016.

\bibitem{meanteacher}
Antti Tarvainen and Harri Valpola.
\newblock Mean teachers are better role models: Weight-averaged consistency
  targets improve semi-supervised deep learning results.
\newblock In {\em Advances in neural information processing systems}, pages
  1195--1204, 2017.

\bibitem{touvron2019fixingresolution}
Hugo Touvron, Andrea Vedaldi, Matthijs Douze, and Herv{\'e} J{\'e}gou.
\newblock Fixing the train-test resolution discrepancy.
\newblock In {\em Advances in Neural Information Processing Systems}, pages
  8252--8262, 2019.

\bibitem{hrnet}
Jingdong Wang, Ke Sun, Tianheng Cheng, Borui Jiang, Chaorui Deng, Yang Zhao,
  Dong Liu, Yadong Mu, Mingkui Tan, Xinggang Wang, et~al.
\newblock Deep high-resolution representation learning for visual recognition.
\newblock {\em IEEE transactions on pattern analysis and machine intelligence},
  2020.

\bibitem{xiao2010sun}
Jianxiong Xiao, James Hays, Krista~A Ehinger, Aude Oliva, and Antonio Torralba.
\newblock Sun database: Large-scale scene recognition from abbey to zoo.
\newblock In {\em 2010 IEEE computer society conference on computer vision and
  pattern recognition}, pages 3485--3492. IEEE, 2010.

\bibitem{xie2019uda}
Qizhe Xie, Zihang Dai, Eduard Hovy, Minh-Thang Luong, and Quoc~V Le.
\newblock Unsupervised data augmentation for consistency training.
\newblock {\em arXiv preprint arXiv:1904.12848}, 2019.

\bibitem{xie2020noisestudent}
Qizhe Xie, Minh-Thang Luong, Eduard Hovy, and Quoc~V Le.
\newblock Self-training with noisy student improves imagenet classification.
\newblock In {\em Proceedings of the IEEE/CVF Conference on Computer Vision and
  Pattern Recognition}, pages 10687--10698, 2020.

\bibitem{resnext}
Saining Xie, Ross Girshick, Piotr Doll{\'a}r, Zhuowen Tu, and Kaiming He.
\newblock Aggregated residual transformations for deep neural networks.
\newblock In {\em Proceedings of the IEEE conference on computer vision and
  pattern recognition}, pages 1492--1500, 2017.

\bibitem{yalniz2019billion}
I~Zeki Yalniz, Herv{\'e} J{\'e}gou, Kan Chen, Manohar Paluri, and Dhruv
  Mahajan.
\newblock Billion-scale semi-supervised learning for image classification.
\newblock {\em arXiv preprint arXiv:1905.00546}, 2019.

\bibitem{yim2017fsp}
Junho Yim, Donggyu Joo, Jihoon Bae, and Junmo Kim.
\newblock A gift from knowledge distillation: Fast optimization, network
  minimization and transfer learning.
\newblock In {\em Proceedings of the IEEE Conference on Computer Vision and
  Pattern Recognition}, pages 4133--4141, 2017.

\bibitem{yuan2019kd-labelsmooth}
Li Yuan, Francis~EH Tay, Guilin Li, Tao Wang, and Jiashi Feng.
\newblock Revisit knowledge distillation: a teacher-free framework.
\newblock {\em arXiv preprint arXiv:1909.11723}, 2019.

\bibitem{yuan2019object}
Yuhui Yuan, Xilin Chen, and Jingdong Wang.
\newblock Object-contextual representations for semantic segmentation.
\newblock {\em arXiv preprint arXiv:1909.11065}, 2019.

\bibitem{zhao2017pyramid}
Hengshuang Zhao, Jianping Shi, Xiaojuan Qi, Xiaogang Wang, and Jiaya Jia.
\newblock Pyramid scene parsing network.
\newblock In {\em Proceedings of the IEEE conference on computer vision and
  pattern recognition}, pages 2881--2890, 2017.

\bibitem{zhu2019deformable}
Xizhou Zhu, Han Hu, Stephen Lin, and Jifeng Dai.
\newblock Deformable convnets v2: More deformable, better results.
\newblock In {\em Proceedings of the IEEE Conference on Computer Vision and
  Pattern Recognition}, pages 9308--9316, 2019.

\end{thebibliography}
}

\end{document}